\pdfoutput=1

\documentclass[11pt]{article}

\usepackage[]{emnlp2021}

\usepackage{times}
\usepackage{latexsym}

\usepackage[T1]{fontenc}

\usepackage[utf8]{inputenc}

\usepackage{microtype}

\usepackage[linesnumbered, ruled]{algorithm2e}
\usepackage[T1]{fontenc}
\usepackage{dsfont}

\usepackage{mdframed}
\usepackage{booktabs}
\usepackage{graphicx}

\usepackage[normalem]{ulem}
\usepackage{enumitem}
\usepackage{xcolor}
\usepackage{soul}
\colorlet{soulred}{red!30}
\sethlcolor{soulred}%

\usepackage{amsmath,amsthm,amsfonts,amssymb,bm}
\usepackage{framed}
\usepackage{varioref}
\usepackage{float}
\usepackage{multirow}
\usepackage{dashrule}
\usepackage{url}
\usepackage{pifont}
\title{Attribute Alignment: Controlling Text Generation from Pre-trained Language Models}

\author{Dian Yu\textsuperscript{1}, Zhou Yu\textsuperscript{2}, Kenji Sagae\textsuperscript{1}\\
\textsuperscript{1} University of California, Davis\\
\textsuperscript{2} Columbia University\\
\texttt{\{dianyu, sagae\}@ucdavis.edu}\\
\texttt{zhouyu@cs.columbia.edu}
}

\begin{document}
\maketitle
\begin{abstract}
Large language models benefit from training with a large amount of unlabeled text,
which gives them increasingly fluent and diverse generation capabilities.
However, using these models for text generation that takes into account target attributes, such as 
sentiment polarity or specific topics, remains a challenge.
We propose a simple and flexible method for controlling text generation by aligning disentangled attribute representations. 
In contrast to recent efforts on training a discriminator to perturb the token level distribution for an attribute, we use the same data to learn an alignment function to guide the pre-trained, non-controlled language model to generate texts with the target attribute without changing the original language model parameters.
We evaluate our 
method on sentiment- and topic-controlled generation,
and show large performance gains over previous methods while retaining fluency and diversity.
\end{abstract}

\section{Introduction}
\label{intro}

While large pre-trained language models (LM) have advanced text generation with coherent language by training on a large amount of unlabeled data \cite{gpt, xlnet, t5}, they are not controllable. For instance, given the prompt ``The issue focused on'',  GPT-2 \cite{gpt2} can generate a high-quality sentence, but it cannot take extra input such as ``positive'' or ``business'' to guide the sentence towards a positive sentiment or business-related topic, due to the lack of attribute labels during training.

To solve the discrepancy between training and inference, one direction is to train an LM from scratch with some supervision such as control codes in CTRL \cite{ctrl}. Nevertheless, this method requires training an LM with a large number of parameters, and is limited by the attributes used during pre-training. Another direction is to fine-tune the pre-trained LM on some annotated datasets. This usually requires updating all the parameters in the model, which incurs large computational
costs with current large LMs that have millions or billions of parameters, and may result in an LM highly relevant only to the specific training data. For example, one can fine-tune a large pre-trained LM on product reviews labeled with sentiment to generate positive and negative sentences, but the fine-tuned model will tend to generate sentences like those from product reviews which greatly limits its utility with out-of-domain prompts.
Both these methods require training all the parameters of the model.
Alternatively, recent research leverages a discriminator to re-weight output distributions \cite{holtzman-etal-2018-learning} or to perturb latent representations in the token level such as in PPLM \cite{pplm} without changing the pre-trained LM.
However, raising target-relevant token probabilities may lead to less fluent sentences. In addition, updating gradients at the token level makes decoding expensive and slow.


In this paper, we propose \texttt{Attribute Alignment} to infuse attribute representations into a pre-trained unconditional LM without changing the LM parameters.
We are inspired by language codes which guide multilingual translation models to translate to the target language \cite{google_mt}. However, because attributes signals are not trained with the LM during large-scale pre-training \cite{google_mt, ctrl}, we introduce an alignment function to bridge attribute representations to the LM so that it can interpret the weights in the attribute representations. 


\begin{table*}[h]
 \small
  \centering
  \begin{tabular} {  p{1.5cm} | p{13cm}  } 
    \hline
 \textbf{Attribute} & \textbf{Generated Text}\\
 \hline
 None & \underline{The issue focused on} a 2008 decision by the United States Court of Appeals for the Ninth Circuit, in San Francisco, that denied local restaurants advance notice of changes to their menus, even when that change had not been submitted to ...\\
 \hline
 positive &\underline{The issue focused on} returning to the simple premise that dialogue is more effective than banal reactions. They demonstrate very good personal style with establishing dialogue and bringing about a good point of view. Most fantastic of all ...\\
 \hline
 negative &\underline{The issue focused on} a false belief that treatment can never be "good enough" and that long-term treatment only "cures" a person. This does not account for why this is the case: Patients with the ...\\
 \hline
 business & \underline{The issue focused on} the regulations preventing banks and other entities in the financial sector from moving money across foreign borders without the consent of its investors. \\
 \hline
 athlete & \underline{The issue focused on} Robinson, who went to camp with his hometown team after being released by the Seattle Seahawks, though it was ruled an emergency by the National Football League.\\
 \hline
 military & \underline{The issue focused on} whether servicemen and women should be allowed to opt out of serving overseas. It was also about whether making it easier for American troops to return home would help their families.\\
 \hline
 world + \hspace{0.05cm}    science & \underline{The issue focused on} an allegation that White House chief science adviser, Michael Mann, misstated data about global warming in his\\
 \hline
  \end{tabular}
  \caption{Examples generated using the proposed alignment function with Bayes disentanglement (\texttt{ACB}). Tokens underscored are the prompts. We use a classifier to select sentences (see Section \ref{exp:eval}) with the highest target attribute predication probability and present the examples here (i.e., the results are not cherry-picked). ``None'' indicates non-controlled generation (original GPT-2 model). ``business'' is from AG News, ``athlete'' is from DBpedia corpus, and ``military'' is not in the training data (zero-shot). ``world + science'' controls multiple attributes.}
 \vspace{-1em}
\label{table:control_example}
\end{table*}

Specifically, we encode an \emph{attribute} (e.g. positive, negative, business, military, etc.) with a pre-trained LM and learn an alignment function to transform the attribute representation.
To train the alignment function, we use the same annotated data used to train discriminators in token-level perturbation methods \cite{pplm} so that the self-attention to the aligned attribute representation will guide the LM with a language modeling objective on the attribute-related dataset.
In contrast to fine-tuning, this does not involve training LM parameters, thus we can do controlled text generation without sacrificing the linguistic quality of the original LM. In addition, we disentangle undesirable features from the training data using a principled approach based on Bayes' Rule. 
Because of the way the attributes are encoded, the end result is that
the generation process can be controlled using arbitrary attributes expressed
as words or phrases. Table \ref{table:control_example} shows text generated
using the prompt \emph{The issue focused on} with various control attributes.
We evaluate our proposed method on sentiment and topic control and show better performance than previous state-of-the-art methods in controlling effectiveness and language quality \footnote{Our code is available at \url{https://github.com/DianDYu/attribute_alignment}}. 


\section{Related Work}
\paragraph{Controlled text generation}
To interpolate a controlling factor,
concatenating the attribute to the input sequence is the most straightforward approach and has been commonly used in grounded generation \cite{wow, prabhumoye-etal-2020-exploring}. \citet{ctrl} proposes to pre-train a large conditional language model with available labels such as URLs for large LM control.
This method can be effective in conditional modeling, but requires a substantial amount of resources for pre-training and is limited by the labels used during pre-training (e.g. 55 control codes in CTRL).
Another approach is to concatenate the attribute representation to the hidden states using linear transformation \cite{concat_hidden, style_transfer} or latent variables \cite{vae, tgvae}. These approaches require training from scratch or fine-tuning the entire pre-trained model to incorporate the external target attributes and model conditional probability \cite{conditional_style_transfer, ziegler-etal-2019-fine, smith-etal-2020-controlling}. In addition, they always require carefully designed Kullback-Leibler (KL)-Divergence and adversarial training to generate out-of training domain text with the desirable attribute only \cite{adversarial_style}. In comparison, our proposed method does not require fine-tuning the original LM so that we can make use of the high quality pre-trained LM while controlling the target attributes.

\begin{figure*}[t]
    \centering
     \includegraphics[width=5.5in]{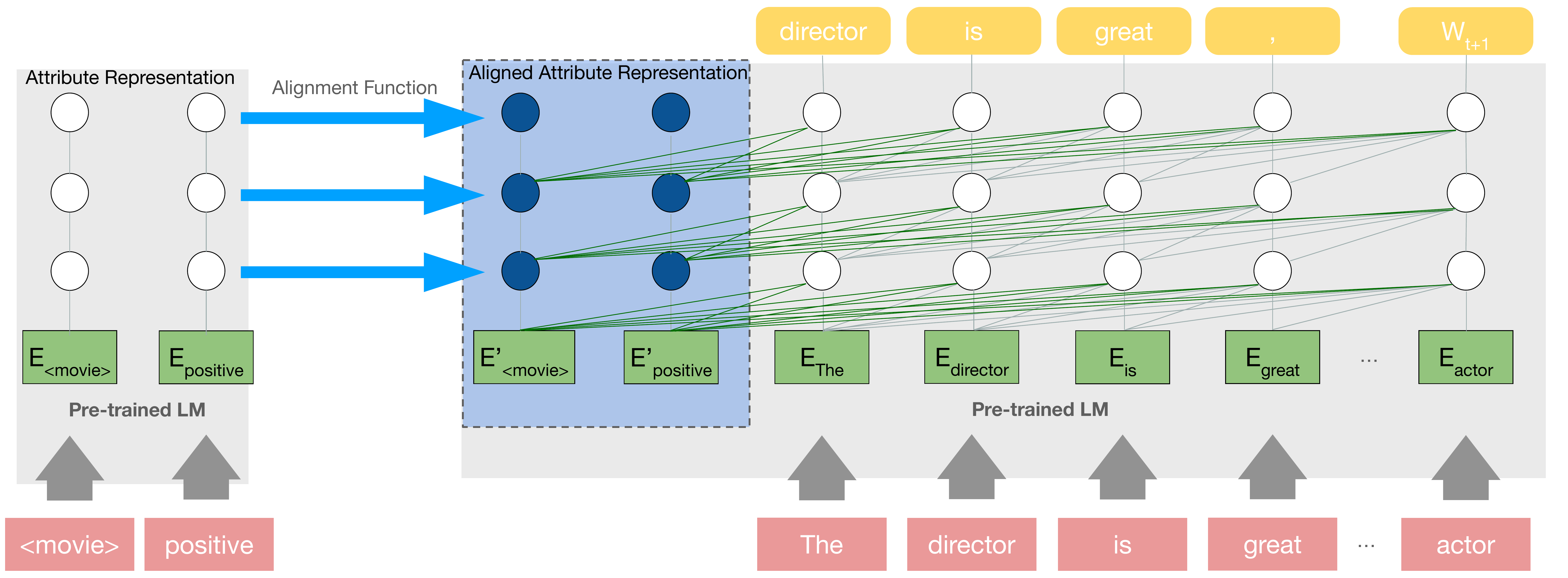}
    \caption{\label{fig:model} 
    Attribute Alignment model architecture with corpus representation disentanglement. We train the alignment function (an MLP in our experiment shown as blue arrows) to transform attribute (e.g. \texttt{positive sentiment}) representation (encoder hidden states in the left grey box) to aligned attribute representation (blue shade box in the middle). The training objective is to generate attribute-related sentences in the training dataset by attending to aligned attribute representation (green lines) in addition to regular self-attention (grey lines).
    }
\end{figure*}

Instead of fine-tuning the whole model, \citet{adapters} proposes to add residual adapters, which are task-specific parameters to transformer layers for each language understanding task. Different from adding adapters for each individual attribute \cite{bapna-firat-2019-simple, ziegler-etal-2019-encoder}, our method only requires learning one attribute alignment function for all attributes to do controlled generation, and is more flexible at inference time without degrading quality such as diversity \cite{madotto-etal-2020-plug}.
Recently, \citet{chan-etal-2021-cocon} proposes to use self-supervised learning with hand-crafted phrases (e.g. ``is perfect'' to represent positive sentiment), but suffers from high variance, low coherence and diversity in order to incorporate the target phrase.
An alternative is to take a pre-trained unconditional LM and perturb the hidden states towards a target attribute in a plug and play manner \cite{plug_and_play}. PPLM proposes to train a classifier or bag-of-words to increase the likelihood of the target attribute in the hidden state for each token \cite{pplm}. Similar to ours, their method does not require changing the pre-trained LM and they are able to control sentiment and various topics. However, ascending conditional probability in the token level to shift the distribution towards target-related tokens 
can lead to degeneration \cite{degeneration} and is slow at inference time. 
The most similar work to ours is probably GeDi \cite{krause-etal-2020-gedi} which proposes to apply weighted decoding using class-conditional LMs with Bayes' Rule on each token to solve the slow inference problem.
Concurrently, \citet{li-etal-2021-prefix} introduces learning prefix rather than task instructions \cite{brown-etal-2020-langauge} and achieves better performances than adapter-based lightweight baselines.
In contrast, our method learns an alignment function on hidden representations of the attribute so that tokens can do self-attention with the attribute without breaking the pre-trained self-attention in the LM. During generation, we can simply send the attribute as a signal for conditional generation. Our method is uniform for different attributes such as sentiment and topics, and is more efficient and flexible.

\paragraph{Attribute representation learning}
\citet{split_representation} splits hidden representations to encourage different dimensions to learn different attributes for document representation. In comparison, \citet{adversarial_style} uses adversarial learning methods to disentangle different attributes such as style. Similarly, \citet{sentiment_neuron} trains a LM on a sentiment classification dataset and finds that one neuron is responsible for the sentiment value in generation. Our proposed disentanglement methods, on the other hand, encourages the alignment function to encode different attributes to different representations and we leverage Bayes' Rule to further separate attributes.

In machine translation, a language representation is learned by appending a language code to the source sentence \cite{google_mt} or summing with word embeddings \cite{xlm} to guide the translation towards the target language. Inspired by these methods \cite{yu-etal-2021-language}, \texttt{Attribute Alignment} appends the attribute to the beginning of a sentence and learns an attribute alignment function to transform attribute representations while freezing the LM parameters, without fine-tuning the whole model in previous methods.

\section{Methodology}
Unconditional language models are trained to optimize the probability of $p(x_{i}|x_{0:i-1})$ where $x_i$ is the next token and $x_{0:i-1}$ are already generated tokens.
For controlled generation, we need to model the conditional distribution $p(x_{i}|x_{0:i-1}, \mathbf{a})$ where $\mathbf{a}$ is the attribute for the model to condition on. To make use of large LMs trained on unlabeled data, we need to infuse the attribute $\mathbf{a}$ into the pre-trained unconditional distribution $p(x_{i}|x_{0:i-1})$. 
We introduce \texttt{Attribute Alignment} to this end. Different from fine-tuning the whole LM, our alignment function is the only trainable component while the pre-trained LM parameters are frozen.

\subsection{Attribute representation with alignment function (\textbf{A})}
\label{model:A}
The high-level idea is to append the attribute token to the beginning of a prompt as a signal so that each token in the sentence can attend to the attribute token.
However, this may break the originally learned sequential dependencies because now the sentence starts with an attribute token followed by a regular sentence, different from the data used for large LM pre-training.

Instead, \texttt{Attribute Alignment} first gets the hidden states of the attribute by running the pre-trained LM on $\mathbf{a}$. Then we align the hidden states using our alignment function ($\mathcal{F}$), implemented as a multi-layer perceptron (MLP) with non-linear connections in this paper,  to get aligned attribute representation. Specifically, in the Transformer architecture \cite{transformer} where hidden states are represented as key-value pairs, the key ($K$) and value ($V$) pair after attribute representation alignment is represented by
\begin{equation}
\label{equ:a}
    K_{:t}', V_{:t}' = [\mathcal{F}(K_\mathbf{a}); K_{:t}], [\mathcal{F}(V_\mathbf{a}); V_{:t}]
\end{equation}
$K_\mathbf{a}$, $V_\mathbf{a}$ are from $LM(x_\mathbf{a})$ and $K_{:t}$, $V_{:t}$ are from $LM(x_{:t})$ where $x_\mathbf{a}$ is the attribute phrase, and $x_{:t}$ are the tokens in the generated sentence up to timestep $t$. Then we can calculate attention and output in the original Transformer model.

During training, we freeze the pre-trained LM and compute the language modeling loss on datasets with the attribute $\mathbf{a}$ to train the alignment function $\mathcal{F}$. The loss function is thus
\begin{equation}
    \mathcal{L}_A = - \sum_{t=0}^{l} \log p(x_t|\mathbf{a}, x_{:t})
\end{equation}
and we only update the parameters of the alignment function using the gradients.
Fig.\ref{fig:model} illustrates the model architecture. 
At inference time, all tokens starting from the prompt attend to the target attribute representation transformed by the trained alignment function in addition to the standard self-attention to generate the next token.
Intuitively, this can be considered as a conditional LM because all tokens now can attend to the aligned attribute representation.

\subsection{Disentangle irrelevant attributes}
The learned alignment function bridges the attribute representation to pre-trained LMs. However, we do not disentangle different features in the training data.
For instance, if we train the alignment function on a movie review dataset for sentiment control, then $\mathcal{F}$ encodes both sentiment and movie review style after aligning the sentiment attribute representation.
Thus, the target attribute representation may be diluted.
To solve this problem, we propose three disentanglement methods. 

\subsubsection{Attribute representation with corpus representation disentanglement (AC)}
We propose to add a corpus domain representation $\mathbf{d}$ along with the attribute representation $\mathbf{a}$ during training. For a training corpus (such as movie reviews) with multiple attributes (such as positive and negative sentiment), $\mathbf{d}$ is used in all the training data while $\mathbf{a}$ is only used in a subset of the training data labeled with the target attribute. Similar to \citet{split_representation}, this can encourage the model to encode target attribute and other features separately into different representations.
Specifically, the key-value pairs can be represented as 
\begin{equation}
\small
    K_{:t}'', V_{:t}'' = [\mathcal{F}(K_\mathbf{a}); \mathcal{F}_\mathbf{d}(K_\mathbf{d}); K_{:t}], [\mathcal{F}(V_\mathbf{a}); \mathcal{F}_\mathbf{d}(V_\mathbf{d}); V_{:t}]
\end{equation}
where $\mathcal{F}_\mathbf{d}$ is a separate alignment function for corpus domain representation, and $K_\mathbf{d}$, $V_\mathbf{d}$ are from the LM encoding of corpus domain names. 
Compared to attributes, corpus domain names might be more abstract so we use special tokens for $\mathbf{d}$ (such as \texttt{<movie review>}) 
and the original texts for attributes (such as \texttt{athlete}). At inference time, we want to generate coherent sentences given any (including out-of-domain) prompts. Therefore, we ignore the corpus representation while having tokens attend to the attribute representation in addition to normal self-attention as in Equation \ref{equ:a} \footnote{In other words, if corpus representation is considered, generating movie reviews or wikipedia-type sentences for any prompt will greatly limit its utility}. 

\subsubsection{KL disentanglement (ACK)}
We also experiment with adding KL-Divergence on top of AC to ensure that the LM does not diverge too much from the original distribution when an attribute signal is added following \cite{pplm}. The disadvantage of this method, however, is that KL-Divergence may also prevent the alignment function from learning useful updates to attribute representation.

\subsubsection{Bayes disentanglement (ACB)}
To further disentangle different features, we use Bayes' Rule to split domain-relevant distribution from attribute-relevant distribution. 
Derived from Bayes' Theorem (See Appendix \ref{bayes}), we have
\begin{equation}
    p(x|\mathbf{a}) \sim \frac{p(x|\mathbf{a}, \mathbf{d})}{p(x|\mathbf{d})} \cdot \frac{p(x, \mathbf{a})}{p(\mathbf{a}| x, \mathbf{d})}
\end{equation}
$p(x|\mathbf{a}, \mathbf{d})$ is the probability distribution of the generated sentence conditioning on both the attribute and the corpus domain, while $p(x|\mathbf{d})$ is the probability distribution of the generated sentence conditioning on the corpus domain only. During training, we assume that different attributes in a corpus (e.g. different sentiments in movie reviews) are close to a uniform distribution. Hence, we consider $p(a|x,d)$ as a constant for a given sentence $x$ from the corpus $d$. Likewise, we consider $p(x,a)$ as a probability distribution from the frozen pre-trained LM with roughly comparable attribute distribution on any sentence to approximate $p(a|x)$, similar to \citet{jiwei}. Therefore, we approximate this equation by eliminating the rest where the elimination does not directly impact a specific training sentence for the target conditional distribution.
We can approximate the desired conditional probability in the log space as 
\begin{equation}
    \log p(x|\mathbf{a}) \sim \log p(x|\mathbf{a}, \mathbf{d}) - \log p(x|\mathbf{d})
\end{equation}
During training, we train the attribute and domain alignment functions ($\mathcal{F}, \mathcal{F}_\mathbf{d}$) by running the LM conditioned on both attribute and domain ($p(x|\mathbf{a}, \mathbf{d}$)), and on domain only ($p(x|\mathbf{a}$)).
In specific, the loss function is 
\begin{equation}
\small
    \mathcal{L}_{ACB} = - \sum_{t=0}^{l} \log p(x_t|\mathbf{a}, \mathbf{d}, x_{:t}) + \sum_{t=0}^{l} \log p(x_t|\mathbf{d}, x_{:t})
\end{equation}
Similar to other proposed methods, the loss is used to update $\mathcal{F}$ and $\mathcal{F}_\mathbf{d}$.
At inference time, suggested by \citet{jiwei}, we use a hyper-parameter $\lambda$ to balance the two distributions. Therefore, the distribution we sample tokens from is 
\begin{equation}
    \log p(x|\mathbf{a}) \sim \log p(x|\mathbf{a}, \mathbf{d}) - \lambda \log p(x|\mathbf{d})
\end{equation}

\subsection{Multi-attribute Control and Zero-shot Inference}
We can simply concatenate aligned attribute representations to control multiple attributes at the same time. In addition,
as we learn the alignment function on the attribute hidden representation from word embeddings instead of learning the attribute representation directly \cite{ziegler-etal-2019-encoder}, we can switch in any attribute token at inference time. Therefore, we can choose attributes not seen in the training corpus and generate text conditioned on a new topic as a zero-shot setting.

\section{Experiments}
We evaluate our proposed methods \textbf{A}: using attribute representation only; \textbf{AC}: Model A with corpus representation for disentanglement; \textbf{ACK}: AC with KL disentanglement; and lastly \textbf{ACB}: AC with Bayes disentanglement. We evaluate these models on sentiment control for thorough comparisons. We use nucleus sampling \cite{degeneration} for all the methods at inference time.  Refer to Appendix \ref{reproducibility} for implementation details.

\subsection{Sentiment control}
\label{sentiment_control_experiment}

\textbf{Data.} We use the Stanford Sentiment Treebank (SST, \citealp{sst}) as our training data. We choose the sentences with positive and negative sentiment to train our alignment function. We select the same 15 prompts such as ``Once upon a time'' that were used in prior work, which were originally randomly selected, and are listed in Appendix \ref{appendix:prefixes} \cite{pplm}.

\noindent\textbf{Baselines.} We compare with five baselines.
\textbf{GPT2} generates unconditioned sentences given the prompts from pre-trained GPT2-medium. The generated sentences are coherent and consistent, but may not capture the target attribute. Its fluency, diversity, and how much the results look like a particular training corpus serve as an upper bound. 
\textbf{GPT2-concat} appends the sentiment token (i.e., \texttt{positive}, \texttt{negative}) before the prompt. It shares the same motivation as our model (see Section \ref{model:A}).
\textbf{GPT2-finetune} is GPT2 fine-tuned with all the model parameters on the same SST dataset by appending an attribute token to the beginning of a sentence. 
Its sentiment control score is an upper bound. \textbf{PPLM} perturbs pre-trained LMs to incorporate attributes without fine-tuning the LM parameters. Similar to ours, the recent state-of-the-art \textbf{GeDi} incorporates target attributes by weighted decoding on the token-level and uses Bayes' Rule on all control codes (rather than domain) to remove unwanted attributes.
It serves as a strong baseline.



\subsection{Topic control}
\textbf{Data.} For topic control, we use AG News dataset \cite{topic_dataset} with four topic attributes (``World'', ``Sports'', ``Business'', ``Sci/Tech'') and DBpedia \cite{topic_dataset} with 14 topic attributes such as ``natural place'' (see Appendix \ref{appendix: dbpedia} for the full list)
as our training data. We use the same 20 prompts from \citet{pplm} (see Appendix \ref{appendix:prefixes}). AG News dataset collects news articles whereas DBpedia dataset collects entity definitions from Wikipedia.

\noindent\textbf{Baselines.} 
PPLM uses different methods for topic control (pre-defined bag of words). For fair comparison, we only compare with \textbf{GPT2}, \textbf{GPT2-finetune}, and \textbf{GeDi} training on the same data. We choose the best preforming models from sentiment control for topic control experiments (\textbf{AC}, \textbf{ACB}), while having ablation study among proposed models on sentiment control.

\subsection{Evaluation}
We evaluate our proposed methods and baselines on sentiment and topic control.
Following \citet{pplm}, we sample ten sentences in a batch and select the most attribute-relevant one over three runs for human evaluation for each prompt in each target attribute. For automatic evaluation, we compare the average performance on all the 30 ($3 \times 10$) conditionally generated results to test the average performance and stability against variances.

\subsubsection{Automatic evaluation}
\label{exp:eval}
We evaluate the conditional generation results on fluency, diversity, attribute relevance, and training data corpus resemblance. 


\noindent\textbf{Fluency} is measured by GPT2-large, a pre-trained external LM, different from the LM we conduct our experiments with (GPT2-medium). We get the average perplexity of the generated sentences (including the prepended prompt).
The perplexity score also indicates how much the generated examples diverge from the pre-trained LM. 

\noindent\textbf{Diversity} is measured by distinct uni-, bi-, and tri-gram ratios as Dist-1, Dist-2, and Dist-3 \cite{jiwei} averaged over all generated sentences.

\noindent\textbf{Attribute relevance} measures how well the generated examples condition on the target attributes. We train classifiers to predict the probability that a given sentence has the target attribute.
For sentiment control, we train an external sentiment classifier using IMDB movie review dataset \cite{imdb} with a BERT \cite{bert} classifier. The classifier achieves an accuracy of $88.51\%$ on the IMDB test set. We also experiment with an internal sentiment classifier trained with SST development set, and we observe that the prediction on the generated texts is similar to that with the external classifier. 

For topic control, we train multi-class classifiers with BERT using $80\%$ of the development sets of AG News and DBpedia datasets. The classifiers achieve an accuracy of $89.71\%$ and $99.25\%$ on the rest of the two development sets, respectively. 
Because other datasets do not share the same topics, we cannot train external classifiers.


\noindent\textbf{Training data corpus resemblance} is used to evaluate if the proposed methods generate sentences that contain undesirable features such as style from the training corpus. For instance, because our proposed method trains with a movie review dataset, the generated examples may tend to be semantically similar to movie reviews.
Similar to attribute relevance, we train a BERT classifier 
by randomly selecting 2,000 training examples and 500 development examples from each of SST, DBpedia, and AG News, and the trained classifier achieves an accuracy of $99.3\%$.
We report the probability that a generated sentence is from its controlling attribute training corpus as the corpus resemblance score.

\begin{table*}[t]
\begin{center}
\resizebox{\textwidth}{!}{
\begin{tabular}{l|cc|ccccc|cc}
\toprule
 \multirow{3}{*}{Model} & \multicolumn{2}{c|}{Attribute} & \multicolumn{5}{c|}{Quality} & \multicolumn{2}{c}{Data} \\
\cmidrule(lr){2-4} \cmidrule(lr){5-8} \cmidrule(lr){9-10}

 & Sentiment & Sentiment & PPL  & Dist-1  & Dist-2  & Dist-3 & Quality & Corpus resemblance & Corpus resemblance\\
 & (classifier) $\%\uparrow$ & (human) $\%\uparrow$ & $\downarrow$ & $\uparrow$ & $\uparrow$ & $\uparrow$ & (human) $\uparrow$ & (classifier) $\%\downarrow$ & (human) $\%\downarrow$ \\
   \midrule
   \multicolumn{4}{l}{\textit{Baselines}}     \\
   \midrule
    GPT2 & 49.24 & - & 37.78 & \textbf{0.49} & \textbf{0.85} & \textbf{0.91} & - &  \textbf{18.31} & -\\
    GPT2-concat & 52.24	& -	& 57.50	& 0.49	& 0.84 &	0.89 & - & 18.87 & -\\

    PPLM & 57.03 & - & 54.03 & 0.44 & 0.79 & 0.88 & - & 26.12 & -\\
    GeDi & 40.03 & 2.18 & 63.49  & 0.36 & 0.77 & 0.86 & 2.91 & 26.31 & 1.44\\
    \midrule
    \multicolumn{4}{l}{\textit{Attribute Alignment}}     \\
    \midrule
    A & 52.61 & - & 40.19 & 0.45 & 0.82 & 0.90 & - & 59.13 & - \\
    AC & 68.92 & - & 48.78 & 0.47 & 0.84 & \textbf{0.91} & - & 62.13 & -\\ 
    ACK & 64.89 & - & 52.66 & 0.48 & 0.84 & \textbf{0.91} & - & 62.80 & -\\
    
   ACB & 64.49 & \textbf{3.49} & 36.62 & 0.48 & \textbf{0.85} & \textbf{0.91} & \textbf{3.25} & 24.05 & 1.91\\
   \midrule
   \multicolumn{4}{l}{\textit{Language model fine-tuning}}     \\
   \midrule
    GPT2-finetune & \textbf{78.78} & -  & 55.60 & 0.37 & 0.66 & 0.75 & - & 92.24 & -\\

\bottomrule
\end{tabular}

}
\end{center}
\caption{Results on sentiment control. 
Sentiment relevance, language quality, and corpus resemblance scores evaluated by humans are in scale of 1-5.
Our proposed model with Bayes disentanglement (ACB) achieves good performance on sentiment controlling while maintaining high quality language generation. 
Note that even though GPT2-finetune achieves the best sentiment controlling score by training the whole LM, it suffers in generation quality and the generated sentences read like movie reviews.}
\vspace{-1em}
\label{table:sent_results}
\end{table*}

\subsubsection{Human evaluation}
We evaluate the generated sentences on attribute relevance, language quality, and training data corpus resemblance. All the metrics are on 1-5 Likert scale. \textbf{Attribute relevance} and \textbf{Corpus resemblance} are similar to the automatic metrics, measuring the degree to which the generated sentences are relevant to the target attributes, and how much the generated sentences read like from their corresponding training corpus, respectively. Since one can easily increase attribute relevance score by sampling target-related tokens more frequently regardless of coherence and the context, \textbf{Language quality} measures if the generated sentences are coherent, in addition to fluency.
Since GeDi outperforms previous strong baselines including PPLM from both automatic and human evaluation \cite{krause-etal-2020-gedi}, we only do human evaluation comparing our best performing model (ACB) with GeDi.

\section{Results and Analysis}
We show controlled examples in Table \ref{table:control_example} and analyze sentiment and topic control results as follows.
\subsection{Sentiment control}
\textbf{Comparison with baselines.} Table \ref{table:sent_results} shows results on sentiment control. Compared to the pre-trained LM (GPT2, $49.24\%$), all our proposed methods achieve better sentiment controlling scores with a large margin and get similar distinct scores. This shows that our proposed method is effective in sentiment control. 

Even though GPT2-finetune achieves the highest sentiment score ($78.78\%$), it gets higher perplexity, lower distinct scores, and very high corpus resemblance ($92.24\%$).
This implies that we can fine-tune a pre-trained LM to condition on the target attribute but suffer from the cost of being restricted to generating sentences resembling the training data as motivated by Section \ref{intro}. 

All our methods outperform PPLM and GeDi with better sentiment control and diversity while having higher language quality. 
For qualitative comparisons between our proposed method and PPLM, we use the IMDB classifier to rank the most negative sentence generated from 30 examples for each prompt and show the generated results in Appendix \ref{comparison}. Compared to our models, 
PPLM suffers from repetition and degeneration problems suggested by both distinct scores and qualitative analysis from the generated examples.
Similarly, even though GeDi can successfully generate sentiment relevant sentences with prompts similar to the training data (such as ``The book'' for book reviews, \citealp{krause-etal-2020-gedi}), it does not generate coherent examples with target sentiment (2.18 from human annotation) on a more diverse set of prompts.
In contrast, using the aligned attribute representation as a control signal to guide the text generation leads to higher sentiment controlling probabilities while keeping the original quality.

\begin{table*}[t]
\begin{center}
\resizebox{\textwidth}{!}{
\begin{tabular}{l|c|cc|ccccc|c}
\toprule
\multirow{3}{*}{Topic source} &  \multirow{3}{*}{Model} & \multicolumn{2}{c|}{Attribute} & \multicolumn{5}{c|}{Quality} & Data \\
\cmidrule(lr){3-4} \cmidrule(lr){5-9} \cmidrule(lr){10-10}

 & & Relevance  &  Relevance &  Perplexity & Dist-1  & Dist-2 & Dist-3  & Quality  & Corpus resemblance \\
 & & (classifier) $\%\uparrow$ & (human) $\%\uparrow$ & $\downarrow$ & $\uparrow$ & $\uparrow$ & $\uparrow$ & (human) $\uparrow$ & (human) $\%\downarrow$  \\
    \midrule
   
   \multirow{3}{*}{AG News} & GPT2 & 25.43 & - &  38.00 & \textbf{0.49} & \textbf{0.84} & \textbf{0.90} & - & -\\
   & GeDi & \textbf{91.61} & 4.75 & 41.42 & 0.28 & 0.73 & 0.86 & 3.68 & 2.61 \\
    \cmidrule(lr){2-10}
   
   &  AC & 63.38 & - & 32.37 & 0.47 & 0.83 & \textbf{0.90} & - & -\\
   & ACB & 64.80 & 4.54 & \textbf{31.22} & 0.46  & 0.83 & \textbf{0.90} & 3.62 & \textbf{2.47}\\
   
    \midrule
    \midrule
    \multirow{3}{*}{DBpedia}
   &  GPT2 & 6.63 & - & \textbf{37.40} & 0.49 & \textbf{0.84} & \textbf{0.90} & - & -\\
    \cmidrule(lr){2-10}
   &   AC & \textbf{32.98} & - & 60.22 & \textbf{0.50} & \textbf{0.84} & \textbf{0.90} & - & - \\ 
   &   ACB & 32.18 & - & 49.85 & 0.49 & 0.83 & \textbf{0.90} & - & -\\
   
\bottomrule
\end{tabular}
}
\end{center}
\caption{Topic control results with topics from AG News and DBpedia.
Attribute relevance score from human annotation and language quality are in scale of 1-5. 
Our proposed methods outperform the GPT2 baselines by a large margin and achieve similar performance with the state-of-the-art GeDi while having higher diversity scores.
}
\vspace{-1em}
\label{table:topic_results}
\end{table*}

\noindent\textbf{Comparison among proposed methods.} The worse performance of having attribute representation only ($52.61\%$) indicates that the entangled attributes dilute the conditional distribution and result in texts using similar vocabularies suggested by low diversity scores. 
In comparison, adding a corpus representation to disentangle target attributes leads to the best performance on sentiment probability prediction.  Further disentanglement by adding KL-Divergence and separating corpus distribution with Bayes' theorem helps to reach lower perplexity and higher distinct scores as expected, but it hurts the attribute controlling performances. This may be caused by that the attribute and corpus representations in fact still mingle with each other so that when we remove the corpus distribution, we also remove some of the target attribute distribution. 
We also note that without Bayes disentanglement, all the other proposed methods reach much higher training corpus resemblance score (e.g. $62.13\%$ with AC) but still much lower than that from fine-tuning ($92.24\%$). This may be partially explained by that sentences with a strong sentiment are more similar to movie reviews than others
from the training corpus resemblance classifier.
Combining all the metrics, it shows that there is trade-off between sentiment control and generation quality. However, we can still control the sentiment better without the cost of perplexity, diversity, and style convergence than the strong baselines. 

\noindent\textbf{Adversarial prompts results.}
Following \citet{pplm}, we also experiment with generating a sentence to an opposing sentiment from a highly polarized prompt. For example, the goal is to generate a positive sentence with the negative prompt ``The food is awful''. 
Using the external classifier to select the generated examples with the most likely target sentiment, we can obtain sentences such as ``\underline{The food is awful} but the service is amazing!'' which is coherent compared to methods like PPLM and GeDi perturbing on the token level. 
Despite the prompts being very polarized, our method can still lead the text generation to the target sentiment without compromising fluency and diversity.
More importantly, although we train our alignment function in the movie review domain, our generated sentences are not biased towards the domain.
We show comparisons to PPLM and GeDi in Table \ref{appendix:oppo_pplm} in the appendix.


\noindent\textbf{Attribute data influence results.}
\begin{table*}[h]
\begin{center}
\resizebox{\textwidth}{!}{
\begin{tabular}{l|ccc|cccc|c}
\toprule
\multirow{2}{*}{Model} & \multicolumn{3}{c|}{Attribute} & \multicolumn{4}{c|}{Quality} & Data \\
\cmidrule(lr){2-4} \cmidrule(lr){5-8} \cmidrule(lr){9-9}

 & Sentiment$\%\uparrow$ & Positive$\%\uparrow$ & Negative$\%\uparrow$ & PPL$\downarrow$  & Dist-1 $\uparrow$ & Dist-2 $\uparrow$ & Dist-3 $\uparrow$  & Corpus resemblance $\%\downarrow$\\
   \midrule
   AC-S & 67.04 & 81.62 & 54.45 & 38.46 & 0.45 & 0.80 & 0.88 & 63.21\\
    ACB-S & 58.85  & 80.88 & 36.82 & \textbf{33.33} & 0.46 & 0.83 & 0.89 & 28.12\\
\bottomrule
\end{tabular}

}
\end{center}
\caption{Results on sentiment control comparing strong polarized training data.}
\vspace{-1em}
\label{table:sent_polarized_results}
\end{table*}
To evaluate how much attribute relevance in training data influences controlling effect, we experiment with training on strong polarized examples labeled as ``very positive'' and ``very negative'' from SST. We denote the corresponding models as \textbf{AC-S}: AC with strong polarized training data; and \textbf{ACB-S}: AC-S with Bayes disentanglement. Table \ref{table:sent_polarized_results} shows that training with strong polarized data achieves similar controlling ability but suffers from lower diversity. This suggests that our proposed method is not sensitive to the attribute quality in the training data, showing the potential to use less strictly annotated data for controlling more diverse attributes.

\subsection{Topic control}

\textbf{Comparison among different methods.} We present our results on topic control in Table \ref{table:topic_results}. Similar to sentiment control, we observe that our proposed methods significantly outperform the baseline in target topic controlling while holding similar perplexity and distinct scores. Even though the topic relevance score is lower than GeDi from automatic evaluation, ACB performs similarly measured by human annotation in terms of both relevance and language quality, while being much more diverse. In addition, using Bayes' disentanglement results in lower perplexity.
However, compared to sentiment control, further disentanglement derives controlling effect on par with the simple disentanglement ($+1.42\%$ and $-0.80\%$ relative change for AG News and DBpedia)
and generates comparable distinct scores. 
This indicates that topic attribute representations may be less entangled with other features such as style from the training corpus compared to that for sentiment representation. We show analysis of GPT-finetune in Appendix \ref{appendix: topic_ft}.

\noindent\textbf{Comparison between training dataset.} To compare the results between topics from AP News and DBpedia, the perplexity is higher than the baseline and the relative corpus resemblance score is also high for DBpedia. 
We conjecture that this is caused by that topics such as ``educational institution'' may be difficult to associate with prompts such as ``Emphasised are'' in the pre-trained LM. When we control the model to generate sentences with the corresponding attributes, the generation diverges from the pre-trained LM more. However, distinct n-grams are not sacrificed.



\subsection{Comparison to GeDi}
From both sentiment control and topic control, we can see that our propose method is on par or better than GeDi in terms of attribute relevance and language quality, while being much more diverse (more than 10\% averaged absolute points on distinct scores). Qualitatively, because GeDi applies weighted decoding on the token level similar to PPLM, we observe that it indeed boosts attribute-relevant token distribution which may lead to incoherent sentences (such as repeating the same phrase). For instance, regardless of the prompt, country and names (e.g. ``Palestinian'') are frequently sampled for the attribute ``world''. This can be further justified by their lower diversity score compared to the baselines. In addition, since GeDi utilized Bayes' Rule on all attribute codes (in comparison to ours on domains), it can also explain the lower performance on sentiment control where attributes are less decoupled.

\subsection{Multi-attribute control and zero-shot analysis}
\label{appendix: zero_shot}
In Table \ref{table:control_example}, we show examples with controlling multiple attribute (e.g. ``world + science technology"). In addition,
topics such as ``military'' are not in the topic control training corpus
so that they are considered as zero-shot attributes. Our trained alignment function can map unseen attribute representation to the target representation to generate fluent and on-topic sentences. However, this zero-shot ability largely depends on the unseen attribute and the provided prompt. 
Following previous research \cite{ctrl} where there may not be good evaluation metrics for the much harder multi-attribute and zero-shot inference task, we only show generated examples here with limited human annotation results showing better controlling and language quality compared to previous work \cite{ krause-etal-2020-gedi}. We conjecture that our better performance is due to our more flexible alignment structure. In comparison, it is more complicated to compute the contrastive generation decoding method using Bayes rule suggested by \citet{krause-etal-2020-gedi} with more control codes without compromising the marginal distribution.


\section{Conclusion}
In this paper, we propose a simple but effective attribute alignment model for conditional language generation on top of non-controlled pre-trained LM without fine-tuning LM parameters.
We also introduce disentanglement methods to separate different features from the training corpus to further preserve the original pre-trained LM distribution.
Evaluated on sentiment and topic control, we show that our proposed method outperforms the previous methods on attribute control while maintaining language generation quality. 
For future work, we plan to apply the proposed methods on other attributes such as dialog act and explore few-shot learning settings of the training corpus. 

\section*{Acknowledgments}
We thank our anonymous reviewers for constructive suggestions. This work was supported by the National Science Foundation under Grant No. 1840191. Any opinions, findings, and conclusions or recommendations expressed are those of the authors and do not necessarily reflect the views of the NSF.

\section*{Ethical Considerations}
The proposed method is intended to explore approaches to perturb pre-trained large language models. We hope that our method can inspire future research on conditional generation while maintaining the original LM generation quality. Meanwhile, we note that our method can be used to generate negative sentences which may harm some use cases. However, similar to previous research, we can apply our method to control the generation to less toxic directions and reduce the risks of misuse. In addition, our experiments are done on English data, but our method can be applied to any language. We did experiments with the same setting and same data with previous research when we claim better performance.







\bibliography{anthology,custom}
\bibliographystyle{acl_natbib}

\clearpage
\appendix

\section{Appendices}
\label{sec:appendix}
\subsection{Bayes Theorem Proof}
\label{bayes}
By Bayes' Theorem,
\begin{align}
    p(x|a, d) & = \frac{p(x, a, d)}{p(a, d)} \\
    & = \frac{p(d|x, a) \cdot p(x, a)}{p(a, d)} \\
    & = \frac{p(d|x, a) \cdot p(x|a) \cdot p(a)}{p(a, d)} \\
    & = \frac{p(d|x, a) \cdot p(x|a) \cdot p(a)}{p(d|a) \cdot p(a)} \\
    & = \frac{p(d|x, a) \cdot p(x|a)}{p(d|a)} \\
\end{align}
so that we can get 
\begin{align}
    p(x|a) = \frac{p(x|a, d) \cdot p(d|a)}{p(d|x, a)}
\end{align}
Transforming the denominator by 
\begin{align}
    p(d|x, a) & = \frac{p(x | d) \cdot p(d) \cdot p(a|x, d)}{p(x, a)} \\
    & \propto \frac{p(x|d) \cdot p(a|x, d)}{p(x, a)} \\
\end{align}
we can get
\begin{equation}
\begin{split}
     p(x|a) & \sim \frac{p(x|a, d)}{p(x|d)} \cdot \frac{p(x, a)}{p(a| x, d)} \cdot \frac{p(d|a)}{p(d|x, a)} \\
     & \sim \frac{p(x|a, d)}{p(x|d)} \cdot \frac{p(x, a)}{p(a| x, d)}
\end{split}
\end{equation}
It is worth noting that our training and loss are novel and different from the pointwise mutual information proposed in \citet{jiwei}, although the ACB inference equation looks similar. From our training data, we can only model $p(x|a, d)$ but not $p(x|a)$ directly. Here $a$ is the target attribute and $d$ represents domain-relevant(noisy) attributes. Therefore, we propose a detailed method adopting Bayes rules to approximate conditional probabilities $p(x|a)$ by removing $d$ from $p(x|a, d)$. In comparison, \citet{jiwei}'s optimization only models $P(T|S)$ and $P(T)$ without any additional attributes or approximation, where $T$ and $S$ are target and source text in conversations. Therefore, the derivation is different. 

\subsection{Prompts for Experiment}
\label{appendix:prefixes}
We use the same 15 prompts used for sentiment control experiment and 20 prompts used for topic controlling experiment from PPLM \cite{pplm}.

\textbf{Sentiment control:} ``Once upon a time", ``The book", ``The chicken", ``The city", ``The country", ``The horse", ``The lake", ``The last time", ``The movie", ``The painting", ``The pizza", ``The potato", ``The president of the country", ``The road", and ``The year is 1910.". 

\textbf{Topic control:} ``In summary", ``This essay discusses", ``Views on", ``The connection", ``Foundational to this is", ``To review,", ``In brief,", ``An illustration of", ``Furthermore,", ``The central theme", ``To conclude,", ``The key aspect", ``Prior to this", ``Emphasised are", ``To summarise", ``The relationship", ``More importantly,", ``It has been shown", ``The issue focused on", ``In this essay".

\subsection{DBpedia topics}
\label{appendix: dbpedia}
The 14 topics from the DBpedia dataset are: "company", "educational institution", "artist", "athlete", "officeholder", "means of transportation", "building", "natural place", "village", "animal", "plant", "album", "film", "written work". \cite{topic_dataset}

\subsection{Implementation Details}
\label{reproducibility}
We use GPT2-medium \cite{gpt2} with 355M parameters as our pre-trained language model, and GPT2-large with 774M parameters as an external language model to evaluate perplexity. Our implementation is based on an efficient transformer architecture \cite{wolf-etal-2020-transformers} where hidden states are stored as key-value pairs. We implement the alignment function with a multi-layer perceptron (MLP) of two linear layers and a non-linear activation function (ReLU). Both at training and inference time, all tokens in the sentence can attend to the attribute representations as if they are appended to the beginning of the sentence, but we fix the position ids of the sentence to start with 0. We apply nucleus sampling \cite{degeneration} with $p$ set to $0.9$ and generate texts with a maximum length of $40$ for all the experiments.

We did not do exhaustive hyperparameter search. For $\lambda$ used in \textbf{ACB}, we tried $0.1, 0.5, 1$. We choose the best hyperparameters on a held-out set of prompts using the evaluate metrics. We set  $\lambda = 0.1$ and report the results in the paper. Similarly, we experimented with $0.01, 0.1, 1$ for the KL scale and show results in the paper with KL scale set to $0.01$. However, we use the suggested hyperparemters from the paper and code for the baselines we compare with \cite{pplm, krause-etal-2020-gedi}.  On SST training set for sentiment control, each epoch takes about 250 seconds, 720 seconds, 720 seconds, and  720 seconds for \textbf{A}, \textbf{AC}, \textbf{ACK}, and \textbf{ACB} respectively on a RTX 2080 Ti GPU machine. We train for 50 iterations for each model. It takes about 3.5 seconds to generate 30 examples for each prompt with evaluation on proposed evaluation metrics.

In addition, we note that PPLM uses top-k sampling and the sampling method may result in different performance. To eliminate the influence from sampling methods, we also compare our methods with PPLM by top-k sampling and our methods show higher sentiment probability and lower perplexity with the same trend (see Table~\ref{tab:topk} in Appendix \ref{Appendix:topk}).

\paragraph{Computational cost}
Our method requires fewer training epochs, less data, and minimal storage. 
Specifically, it takes fine-tune model 10 epochs (1846.6s), our methods AC 7 epochs (1237.6s), and ACB 7 epochs (1622.6s) during training. It takes 3.4s, 3.4s, 3.5s respectively at inference time to generate 30 examples. Overall, our method is computationally more efficient.  Moreover, recent research suggests that similar alignment methods as ours require less training data than fine-tuning \cite{li-etal-2021-prefix}. With less data, our method would require even less iterations to converge. Additionally, we only need to store the trained alignment function for all attributes, compared to all parameters for fine-tuning, and one adapter per attribute in residual adapters \cite{bapna-firat-2019-simple}.

\newpage
\subsection{Comparison using top-k sampling}
\label{Appendix:topk}

\begin{table}[h]
\begin{center}
\resizebox{\columnwidth}{!}{
\begin{tabular}{lcc}
\toprule
\textbf{Model} & \textbf{Sent. prob.$\%\uparrow$} & \textbf{Perplexity$\downarrow$ } \\
\midrule
GPT2 & 49.98 & \textbf{10.94} \\
PPLM & 58.57& 17.52 \\
\midrule
\midrule
AC & \textbf{67.39} & 16.53 \\ 
ACB & 60.54 & 13.35 \\
\bottomrule
\end{tabular}
}
\end{center}
\caption{Comparison on different methods using top-k sampling ($k=10$).}
\label{tab:topk}
\end{table}


\subsection{Performance of GPT-finetuning on topic control.}
\label{appendix: topic_ft}
Table \ref{table:topic_results_ft} shows results for topic control by fine-tuning GPT-2.
Similar to sentiment control, even though we can achieve better topic relevance score, the generated sentences suffer from low language quality and much less diversity, while converging to the training data. This greatly limits its utility (for example, we may want to generate a coherent sentence about nature in scenarios such as conversations, but we do not want to generate anything that reads like Wikipedia or repeating about forests.)

\onecolumn
\begin{table*}[h]
\begin{center}
\resizebox{\textwidth}{!}{
\begin{tabular}{l|c|c|cccc|c}
\toprule
\multirow{2}{*}{Topic source} &  \multirow{2}{*}{Model} & \multicolumn{1}{c|}{Attribute} & \multicolumn{4}{c|}{Quality} & Data \\
\cmidrule(lr){3-3} \cmidrule(lr){4-7} \cmidrule(lr){8-8}

 & & On topic prob. $\%\uparrow$ & Perplexity$\downarrow$ & Dist-1 $\uparrow$ & Dist-2 $\uparrow$ & Dist-3 $\uparrow$ & Corpus resemblance $\%\downarrow$ \\
    \midrule
    AG News & GPT2-finetung & 77.06 & 30.42 & 0.46 & 0.82 & 0.89 & 98.37\\
    DBpeddia & GPT2-fineune & 59.21 & 69.12 & 0.47 & 0.79 & 0.87 & 58.9\\
  
\bottomrule
\end{tabular}
}
\end{center}
\caption{Topic control results with topics by fine-tuning GPT-2 from AG News and DBpedia.}
\vspace{-1em}
\label{table:topic_results_ft}
\end{table*}

\subsection{Comparison to PPLM and GeDi on adversarial prompts}
\begin{table*}[h]
\small
  \centering
  \begin{tabular} { p{1.1cm} | p{1.2cm} | p{12cm}  } 
  \toprule
 \textbf{Model} & \textbf{Target} & \textbf{Generated Text}\\
 \midrule
 \multirow{5}{*}{ACB} 
 & positive &\underline{The food is awful} but the service is amazing! The takeout is amazing! However, for me, a small, cozy restaurant that is a small institution in a small town I'm so glad that they are planning on ...\\
\cmidrule{2-3}
 & negative &\underline{The food is amazing}!! We didn't want to bring it home as the night before, but we can't afford the honey pot cost so we ended up throwing in our own trail mix as well as having ...\\
  \midrule
  \midrule
 \multirow{7}{*}{PPLM} 
 & positive & \underline{The food is awful} 
 but there is also the music, the story and the magic! \textbackslash n \textbackslash n The "Avenged Sevenfold" is a masterfully performed rock musical that will have a strong presence all over the world ... \\
  \cmidrule{2-3}
 & negative & \underline{The food is amazing}\textbackslash n \textbackslash n  It’s also not. \textbackslash n \textbackslash n  It is not the kind of stuff that you would just want to spend your money on for \textbackslash n \textbackslash n I don’t know why, but when I got my second box, it felt like a bad rip off ... \\
 \midrule
 \midrule
 \multirow{5}{*}{GeDi} 
 & positive & \underline{The food is awful} 
 now but awesome! Love this place!!! (10 minutes away from anywhere I can find good food). Great tapas selection and the chef is very friendly! Excellent staff and great ... \\
  \cmidrule{2-3}
 & negative & \underline{The food is amazing} but terrible...\textbackslash n \textbackslash n Why would I buy to test restaurants when I could order online online or drive home in a car instead? ... \\
 \bottomrule
  \end{tabular}
  \caption{Adversarial sentiment control examples compared to PPLM and GeDi generated from the proposed alignment function with Bayes disentanglement (ACB) where the prompt has a strong opposite sentiment. Similar to Table~\ref{table:control_example}, the results here are selected by a sentiment classifier (not cherry-picked). PPLM results are taken from \citet{pplm}}
\label{appendix:oppo_pplm}
\end{table*}

\subsection{Comparison between Attribute Alignment (ACB) and PPLM Examples}
\label{comparison}

\begin{table*}[t]
  \centering
  \begin{tabular} {  p{1cm}  p{1cm}  p{1cm}  p{11.3 cm}  } 
    \hline
 \textbf{Model} & \textbf{pred$\%$} & \textbf{PPL} &\textbf{Generated Text}\\
 \hline
PPLM* & 98.31 & 22.19 & \underline{Once upon a time} \textbackslash n\textbackslash n I made this game for my wife, and she loved it! I have made a wonderful discovery of how to make this very amazing and beautiful looking and beautiful, beautiful, amazing book! I \\
 \hline
 PPLM & 98.39 & 119.54 & \underline{Once upon a time}, in a distant galaxy, a supernova blast destroyed a supernova explosion the losing side ripping apart sScRush UV-3a. A burnt out and rusty mess of garbage spoods the \\
 \hline
 ACB & 99.52 & 42.17 & \underline{Once upon a time}, eBay lists its canceled items. I don't think there is a list of canceled items that I can see here. In the meantime, a bunch of crap, from iPhones (minus their selling center \\
\hline
\hline
 PPLM* & 96.53 & 13.52 & \underline{The city} of Detroit, the country's third-richest and most-populous, is the most violent, most dysfunctional and most pathetic city in American history; that is, if the United States, which \\
 \hline
 PPLM & 99.88 & 158.57 & \underline{The city} might as well have been written by \textbackslash n\textbackslash n "We got into this mess, how could youWhat. and by" (by the night was "O-but of the/-how we" \\
 \hline
  ACB & 98.07 & 31.98 & \underline{The city} is a city of commerce, traffic and construction. In the midst of all this noise and bustle, West Queen West has produced the most monotonous of urban mini-revolutions: no\\
 \hline
 \hline
 PPLM* & 99.46 & 11.67 & \underline{The book} was not the best. I found some of the writing to be slightly clunky and awkward. I liked some, but not enough to buy a copy. \textbackslash n\textbackslash n The plot seemed to be about a \\
 \hline
 PPLM* & 99.84 & 29.16 & \underline{The book} comes out of the ashes of my writing and writing. It was a complete disaster from beginning to end. I had seasoned 250,000 words--at least an hour's per day to write on it\\
 \hline
 ACB & 99.69 & 27.30 & \underline{The book} was poorly written, written off as 'opinion' and poor grammar and punctuation were used extensively. No wonder the author is currently living in a halfway house with his estate. Nevermind the fact \\
 \hline
 \hline
 PPLM* & 83.42 & 21.89 & \underline{The president of the country}'s largest hospital says they are now on a "mission to save lives" and that "the people of the US" should not pay for this. I'm a man. I am the reason the\\
 \hline
 PPLM & 86.46 & 32.13 & \underline{The president of the country}'s largest college was fired from her job for giggling at a Golden Gate crowd as a crowd in San Francisco rallied to help a storm victim. \textbackslash n\textbackslash n However, the New York Times shames the \\
 \hline
 ACB & 97.93 & 13.32 & \underline{The president of the country} isn't in office. The president isn't in office. You can't make an argument based on that kind of stuff. So what's the point of it?\\
 \hline
 \hline
 PPLM* & 69.73 & 21.67 & \underline{The painting} will be on the back \textbackslash n\textbackslash n - "A \textbackslash n\textbackslash n . \textbackslash n\textbackslash n - n\textbackslash n\textbackslash "$<|$endoftext$|>$"I can't be the only one who has the right to be a 'f*ck you\\
 \hline
 PPLM & 99.44 & 119.54 & \underline{The painting} is thought fluff, a very poor, and a shambolic, modern, and bannister-ly, why did you just do that to me, you and your wretched brand of dreadful\\
 \hline
 ACB & 99.75 & 20.76 & \underline{The painting} is one of the worst I've seen in my lifetime... it's so corny and flat. It's such a cheap, offbeat example. It's more shocking than shocking, because you wouldn\\
 \hline
 \hline
 
 \end{tabular}
\end{table*}
 \begin{table*}[t]
  \centering
  \begin{tabular} { p{1cm}  p{1cm}  p{1cm}  p{11.3 cm} } 
    \hline
 \textbf{Model} & \textbf{pred$\%$} & \textbf{PPL} &\textbf{Generated Text}\\
 \hline
 
 PPLM* & 96.92 & 73.73 & \underline{The horse} has no need for any of this. \textbackslash n\textbackslash n ; ; ; : : ; ; ;; ; ;! \# \#!? :? *? :? no ( : the ( ( @ the ( \\
 \hline
 
 PPLM & 97.72 & 94.93 & \underline{The horse} is a wyvern. A wyder is a "rifle". A good shot. Create Chris C, a pretty, brunette, a skinny, bald drone. Just a fat. \\
 \hline
 ACB & 97.39 & 47.43 & \underline{The horse} he's teaching to lick it away at the bar: heck, the economy would be better off if they didn't have one. In fairness, he could certainly have cut some of his cast more slack
 \\
 \hline
 \hline
 PPLM* & 94.61 & 20.66 & \underline{The lake} has long been the center for a long, ugly, and and and and. \textbackslash n\textbackslash n . \textbackslash n\textbackslash n . \textbackslash n\textbackslash n . \textbackslash n\textbackslash n . \textbackslash n\textbackslash n The problem with the problem is I can't find \\
 \hline
 PPLM & 96.21 & 43.00 & \underline{The lake} around Yaffo in south-central Russia in the world's only biodiversity-poor desert was the scene of the worst air quality in Europe, with more than half of the population suffering three different types \\
 \hline
 ACB & 97.39 & 44.15 & \underline{The lake} is not vast enough to accommodate a tight lake liner. \textbackslash n\textbackslash n 1. Looking for catnap materials in the lake \textbackslash n\textbackslash n Finding a catnap bather or two is like asking a family \\
 \hline
 \hline
 PPLM* & 98.72 & 15.78 & \underline{The country}  is in a tailspin with the economy barely growing and the budget deficit rising. \textbackslash n\textbackslash n The government's budget is a failure. A failure for which there's nothing the public will not pay the price \\
 \hline
 PPLM & 97.20 & 32.84 & \underline{The country}'s will merely sit silently on its grave.
\textbackslash n\textbackslash n A federal government miscalculated and the economy is limping back to the roots. \textbackslash n\textbackslash n Just how bad are the latest developments and what do \\
 \hline
 ACB & 94.53 & 38.63 & \underline{The country}  has become too interested in its politics to pay attention to anything else. The top domestic TV stations should say nothing about this conflict or this nation and instead should be focusing on discussing the place of gays and \\
 \hline
 \hline
 PPLM* & 97.83 & 84.65 & \underline{The road} to the White House is an ugly,,,,,,,,,,,... \textbackslash n \textbackslash n A house or a bin is a a a. It's got an awful, unpleasant name.\\
 \hline
 PPLM & 86.51 & 33.87 & \underline{The road} to intensive genetic counseling \textbackslash n\textbackslash n When can an individual apply for a program under "animal welfare" or "conservation"? \textbackslash n\textbackslash n What does "social welfare," "mangle" and "population\\
 \hline
 ACB & 97.58 & 35.38 & \underline{The road} trip to Seattle and Pacific Ocean was plagued with weird mishaps. Our airline plane couldn't land at Seattle airport so a Toyota Hilux truck was hired to drive it, and a lot of the freight\\
 \hline
 \hline
 PPLM* & 99.86 & 8.94 & \underline{The movie} is a total disaster, and it's been a total failure at that. \textbackslash n \textbackslash n I can't say I'm entirely convinced it's been a total failure, but I can assure you I'm not \\
 \hline
 PPLM* & 99.77 & 34.70 & \underline{The movie} is chock-full of nothing good. It is a horrible 3 minute movie. It would've been terrible if I didn't see it when I was through Foety.\\
 \hline
 ACB & 99.60 & 23.77 & \underline{The movie} set/store just makes no sense at all. \textbackslash n \textbackslash n I've only had my phone with me for a week, so I didn't watch it while others did. On day two I caught the \\
 \hline
 \hline
 PPLM* & 97.65 & 25.52 & \underline{The pizza}iolo or specialised freezer version of an Italian classic is no better or worse than a standard hot dog. The good, you just eat it, while the bad, you cook it way, way \\
 \hline
 PPLM & 99.64 & 20.79 & \underline{The pizza} oven. The pan. What a boring, boring job. You put everything in it, right? So there's this wonderful smell in there. But this is the worst part of it. It takes \\
 \hline
 ACB & 99.18 & 25.45 & \underline{The pizza} box is, in a word, a piece of garbage \textbackslash n The first-ever-to-make-it-in-a-Bowl \textbackslash n An over-sucking, over-dram \\
 \hline
 \hline
 
 \end{tabular}
 \end{table*}
 \begin{table*}[t]
  \centering
  \begin{tabular} { p{1cm}  p{1cm}  p{1cm}  p{11.3 cm} } 
    \hline
 \textbf{Model} & \textbf{pred$\%$} & \textbf{PPL} &\textbf{Generated Text}\\
 \hline

 PPLM* & 95.21 & 8.33 & \underline{The last time} New Orleans faced a top-five opponent, it went 4-13. \textbackslash n \textbackslash n The Pelicans have lost six consecutive games. \textbackslash n \textbackslash n The Pelicans have the league's fourth-worst offense.\\
 \hline
 PPLM & 47.17 & 27.44 & \underline{The last time} I wrote about buryable garbage I compared the North American garbage garbage heap to that of "Hell" on Earth, and if you have been following along you have noticed some large differences. "Hell"\\
 \hline
 ACB & 97.10 & 15.95 & \underline{The last time}  we were all excited about Wolf of Wall Street, it was Sam Jackson. It took for us a while for us to realize that this was going to be a terrible idea. \textbackslash n \textbackslash n Stereotype\\
 \hline
 \hline
 PPLM* & 99.15 & 22.28 & \underline{The year is 1910.} A woman, the only child in a small, isolated village, is attacked and is killed.$<|$endoftext$|>$I have no time for the idiotic and disgusting bullshit that is the NFL, and that is just \\ 
 \hline
 PPLM & 96.24 & 39.15 & \underline{The year is 1910.} Colonists on a long-sought-after research mission return to a barren world of dirt and rubble. The expedition discovers a barren, randy device who possesses a hundred-year-old device \\ 
 \hline
 ACB & 99.93 & 17.29 & \underline{The year is 1910.} He's going back home to Paris, where he's an English salesman. He's trying to raise a family and he's having some trouble when his wife returns from an extended vacation. \\ 
 \hline
 \hline
 PPLM* & 99.30 & 14.63 & \underline{The potato}is the world's most widely eaten meat, and its high price is why we eat so much. But is the potato actually the worst meat you'll get? And does a potato really have the worst \\
 \hline
 PPLM & 99.00 & 166.03 & \underline{The potato}, a slender, poorly vascular plant that is a poor choice for many traditional timesaving reasons. Full of nasty things like the inability to remember details where the it is raised, is the sention the\\
 \hline
 ACB & 99.52 & 66.44 & \underline{The potato} seems to be a slow, vomiting, and hungry thing. I have seen it eat its excess of juice and poop and drink in thin streams. Yet, despite this hideous abnormality, it hardly feels\\
 \hline
 \hline
 PPLM* & 98.47 & 18.20 & \underline{The chicken} wing virus was a terrible thing. I mean, really bad. \textbackslash n \textbackslash n The virus, known as "Chicken Wing," was a disease that was devastating to the entire chicken world, killing thousands of chickens \\
 \hline
 PPLM & 95.96 & 26.20 & \underline{The chicken} coop is a great idea for people, but if you are getting pregnant, the plan is not going to work. Hermies, baby and toddlers are at risk. \textbackslash n\textbackslash n Most people would \\
 \hline
 ACB & 99.24 & 39.75 & \underline{The chicken} commercial is packed full of even more bullshit. For the nearly 900th time, Wendy's CEO Joe Noller has made it clear that there is an organization in this country that hates its products, specifically \\
 \\
 \hline
 \end{tabular}
  \caption{Examples from PPLM\cite{pplm} and our proposed method (ACB: attribute and corpus representation with Bayes disentanglement) for each prompt we experiment with. Note that the perplexity is not comparable among different sampling methods. We use top-p sampling for ACB and and PPLM, and top-k sampling for PPLM* because \citet{pplm} suggests top-k in their paper for the best results. We use an external classifier to select the example with the highest negative probability from 30 generated sentences and present the results.}
\label{table:pplm_comparison}
\end{table*}

\end{document}